\documentclass{article} 
\usepackage{iclr2015,times}
\usepackage{hyperref}
\usepackage{url}
\usepackage{amsfonts}
\usepackage{graphicx}
\usepackage{subfigure}
\usepackage{natbib} 
\usepackage{epstopdf}

\iclrfinalcopy

\title{Towards Deep Neural Network Architectures Robust to Adversarial Examples
}

\author{
Shixiang Gu \\
Panasonic Silicon Valley Laboratory \\
Panasonic R\&D Company of America \\
\texttt{shane.gu@us.panasonic.com} \\
\And
Luca Rigazio \\
Panasonic Silicon Valley Laboratory \\
Panasonic R\&D Company of America \\
\texttt{luca.rigazio@us.panasonic.com} \\
}

\begin{document}

\maketitle

\begin{abstract}
Recent work has shown deep neural networks (DNNs) to be highly susceptible to well-designed, small perturbations at the input layer, or so-called adversarial examples. 
Taking images as an example, such distortions are often imperceptible, but can result in 100\% mis-classification for a state of the art DNN.
We study the structure of adversarial examples and explore network topology, pre-processing and training strategies to improve the robustness of DNNs. 
We perform various experiments to assess the removability of adversarial examples by corrupting with additional noise and pre-processing with denoising autoencoders (DAEs). 
We find that DAEs can remove substantial amounts of the adversarial noise. However, when stacking the DAE with the original DNN, the resulting network can again be attacked 
by new adversarial examples with even smaller distortion.
As a solution, we propose Deep Contractive Network, a model with a new end-to-end training procedure that includes a smoothness penalty inspired by the contractive autoencoder (CAE). 
This increases the network robustness to adversarial examples, without a significant performance penalty.
\end{abstract}

\section{Introduction}
\label{intro}

Deep neural networks have recently led to significant improvement in countless areas of machine learning, from speech
recognition to computer vision~\cite{krizhevsky2012imagenet,dahl2012context,taigman2013deepface,zhang2013panda}.
DNNs achieve high performance because deep cascades of nonlinear units allow to generalize non-locally, in data-specific manifolds~\cite{bengio2009learning}.
While this ability to automatically learn non-local generalization priors from data is a strength of DNNs, 
it also creates counter-intuitive properties. In particular ~\cite{szegedy2013intriguing} showed in their seminal paper that one can engineer small perturbations to the input data, 
called adversarial examples, that make an otherwise high-performing DNN misclassify every example.
For image datasets, such perturbations are often imperceptible to the human eye, thus creating potential vulnerabilities when deploying neural networks in real environments. 
As an example, one could envision situations where an attacker having knowledge of the DNN parameters could use adversarial examples
to attack the system and make it fail consistently. Even worse, due to the cross-model, cross-dataset generalization properties of the adversarial examples~\cite{szegedy2013intriguing}
, the attacker might generate adversarial examples from independent models
without full knowledge of the system and still be able to conduct a highly successful attack.
This indicates there is still a significant robustness gap between machine and human perception, despite recent results showing machine vision performance closing
in on human performance~\cite{taigman2013deepface}. 
More formally, the challenge is: can we design and train a deep network that not only generalizes in abstract manifold space to 
achieve good recognition accuracy, but also retains local generalization in the input space?

A main result from ~\cite{szegedy2013intriguing} is that the smoothness assumption that underlies many kernel methods such as Support Vector Machines (SVMs) 
does not hold for deep neural networks trained through backpropagation.
This points to a possible inherent instability in all deterministic, feed-forward neural network architectures.
In practice, SVMs can be used to replace the final softmax layer in classifier neural networks leading to better generalization~\cite{tang2013deep}, 
but applying SVM in the manifold space does not guarantee local generalization in the input space.
Recently, ~\cite{duvenand14avoiding} categorize distributions of deep neural networks through deep Gaussian Process (GP) and show that in stacked architectures,
the capacity of the network captures fewer degrees of freedom as the layers increase. They propose to circumvent this by connecting inputs to every layer of the network. 
Without this trick, the input locality is hardly preserved in higher layers due to the complexity of nonlinear mapping cascades. 

A framework leveraging both approaches is Random Recursive SVM (R$^{2}$SVM) ~\cite{vinyals2012learning}, which recursively solves a SVM
whose input combines input data and outputs from the previous SVM layer, randomly projected to the same dimension as the input data. 
R$^{2}$SVM avoids solving nonconvex optimization by recursively solving a SVM and demonstrates generalization on small datasets. However,  
performance is suboptimal compared to state-of-the-art DNNs, possibly due to lack of end-to-end training~\cite{vinyals2012learning,tang2013deep}.
Another work inspired by the recursive nature of the human perceptual system is Deep Attention Selective Network (dasNet) ~\cite{stollenga2014deep},
which dynamically  fine-tunes the weight of each convolutional filter at recognition time. 
We speculate that the robustness of human perception is due to complex hierarchies and recursions in the wirings of the human brain~\cite{felleman1991distributed,douglas1995recurrent}, 
since recursions provide multiple paths to input data and could retain locality information at multiple levels of representation. 
Such an intuition is also partially supported by the recent state-of-the-art models for object classification and detection involving multi-scale processing~\cite{szegedy2014going}. 
Since modeling such recursions in DNNs is notoriously hard and often relies on additional techniques such as reinforcement learning~\cite{stollenga2014deep,mnih2014recurrent}, 
we will at first investigate explicit inclusion of input generalization as an additional objective for the standard DNN training process. 

It is important to note that the adversarial examples are universal and unavoidable by their definition:
one could always engineer an additive noise at input to make the model misclassify an example,
and it is also a problem in shallow models such as logistic regression~\cite{szegedy2013intriguing}.
The question is how much noise is needed to make the model misclassify an otherwise correct example.
Thus, solving the adversarial examples problem is equivalent to increasing the noticeability of the smallest 
adversarial noise for each example.  

In this paper we investigate new training procedures such that the adversarial examples generated based on ~\cite{szegedy2013intriguing} have higher distortion, 
where distortion is measured by $\frac{1}{n}\sum(x_i'-x_i)^2$ where $x', x \in \mathbb{R}^n$ are the adversarial data and original data respectively. 
First, we investigate the structure of the adversarial examples, and show that contrary to their small distortion it is difficult to recover classification 
performance through additional perturbations,
such as Gaussian additive noises and Gaussian blur. This suggests the size of ``blind-spots'' are in fact relatively large, in input space volume, and locally continuous. 
We also show that adversarial examples are quite similar~\cite{szegedy2013intriguing}, and an autoencoder (AE) trained to denoise adversarial examples from one network generalizes 
well to denoise adversarials generated from different architectures. However, we also found that the AE and the classifier DNN can be stacked and the 
resulting network can again be attacked by creating new,
adversarial examples of even smaller distortion. Because of this, we conclude that ideal architectures should be trained end-to-end and incorporate input 
invariance with respect to the final network output. 
We find that ideas from denoising autoencoder (DAE), contractive autoencoder (CAE), and most recently marginalized denoising autoencoder (mDAE) provide strong 
framework for training neural networks that are robust against adversarial noises~\cite{rifai2011contractive,alain2012regularized,chen2014marginalized}. 
We propose Deep Contractive Networks (DCNs), which incorporate a layer-wise contractive penalty, and show that adversarials generated from such networks 
have significantly higher distortion. 
We believe our initial results could serve as the basis for training more robust neural networks that can only be misdirected by a substantial noise, 
in a way that is more attuned to how human perception performs. 

\section{Framework}
\label{framework}

\subsection{Generating Adversarial Examples}
\label{gen_adv}

We follow the procedure outlined in~\cite{szegedy2013intriguing} for generating adversarial examples from the classifier neural network,
with a slight modification for computational efficiency.
Using the same notation, we denote the classifier by $f : \mathbb{R}^m \longrightarrow \{1...k\}$ and 
its associated continuous loss function by $L : \mathbb{R}^m \times \{1...k\} \longrightarrow \mathbb{R}^+$.
Then, for a given image $x \in \mathbb{R}^m$, whose $m$ pixels normalized to $[0,1]$, and target label $l \in \{1...k\}$, the minimal adversarial noise
$r$ is approximated by optimizing the following problem, given $c > 0$ and subject to $x+r\in [0,1]^m$:
\begin{equation}
\label{eq:adv}
 \textbf{min}_r\ c|r|_2\ +\ L(x + r, l)
\end{equation}
Instead of finding the minimum $c$ through line-search per example, we use constant $c$ during evaluation of a given dataset and model
architecture. We find $c$ such that for a sufficiently large subset of data of size $n$, $\sum_{i=0}^{n-1}\parallel r_i \parallel_2$ is 
minimized, subject to the constraint that the mean prediction error rate of $f(x_i + r_i)$ is greater than 
$e$. $e$ is chosen as $99\%$ throughout the experiments.
Since we are interested in macro-scale evaluation of adversarial examples per dataset and model, we find this setting sufficiently
simplifies and speeds up the procedure while allowing quantitative analysis of the results.

\subsection{Datasets and Model Architectures}
\label{models}
We perform our experiements on the MNIST dataset, using a number of architectures ~\cite{lecun1998mnist,krizhevsky2009learning}. 
Table~\ref{models-table} summarizes experimental settings, baseline error rate, adversarial examples' error rate and average adversarial distortion. 
$L_2$ weight decay is applied with $\lambda = 10^{-3}$, except in convolutional layers. 
For MNIST ConvNet has two convolutional layers, one fully-connected layer, and one softmax layer. 

\begin{table}[t]
\caption{Model architectures, datasets, and baseline error-rates}
\label{models-table}
\begin{center}
\begin{tabular}{| l |l | l | l | l | l | l | }
\hline
Model Name &Dataset &Description &Train err. &Test err. &Adv. err. &Av. distortion\\
\hline  
N100-100-10             &MNIST		&ReLU net
&0\% &1.77\% &99.9\% &0.084\\
N200-200-10             &MNIST		&ReLU net
&0\% &1.65\% &99.9\% &0.087\\
AE400-10		&MNIST		&AE net
&1.0\% &2.0\% &99.6\% &0.098\\
ConvNet		&MNIST		&Conv. net 
&0\% &0.90\% &100\% &0.095\\
\hline
\end{tabular}
\end{center}
\end{table}

\section{Recovering from Adversarial Examples}
\label{recov_adv}

In order to gain insight into the properties of adversarial noises, we explore three pre-processing methods aiming at 
recovering from adversarial noise, as presented in the following sections.

\subsection{Noise Injection}
\label{sec:noises}

\begin{figure}
 \centering
 \begin{subfigure}
  \centering
  \includegraphics[width=0.49\linewidth]{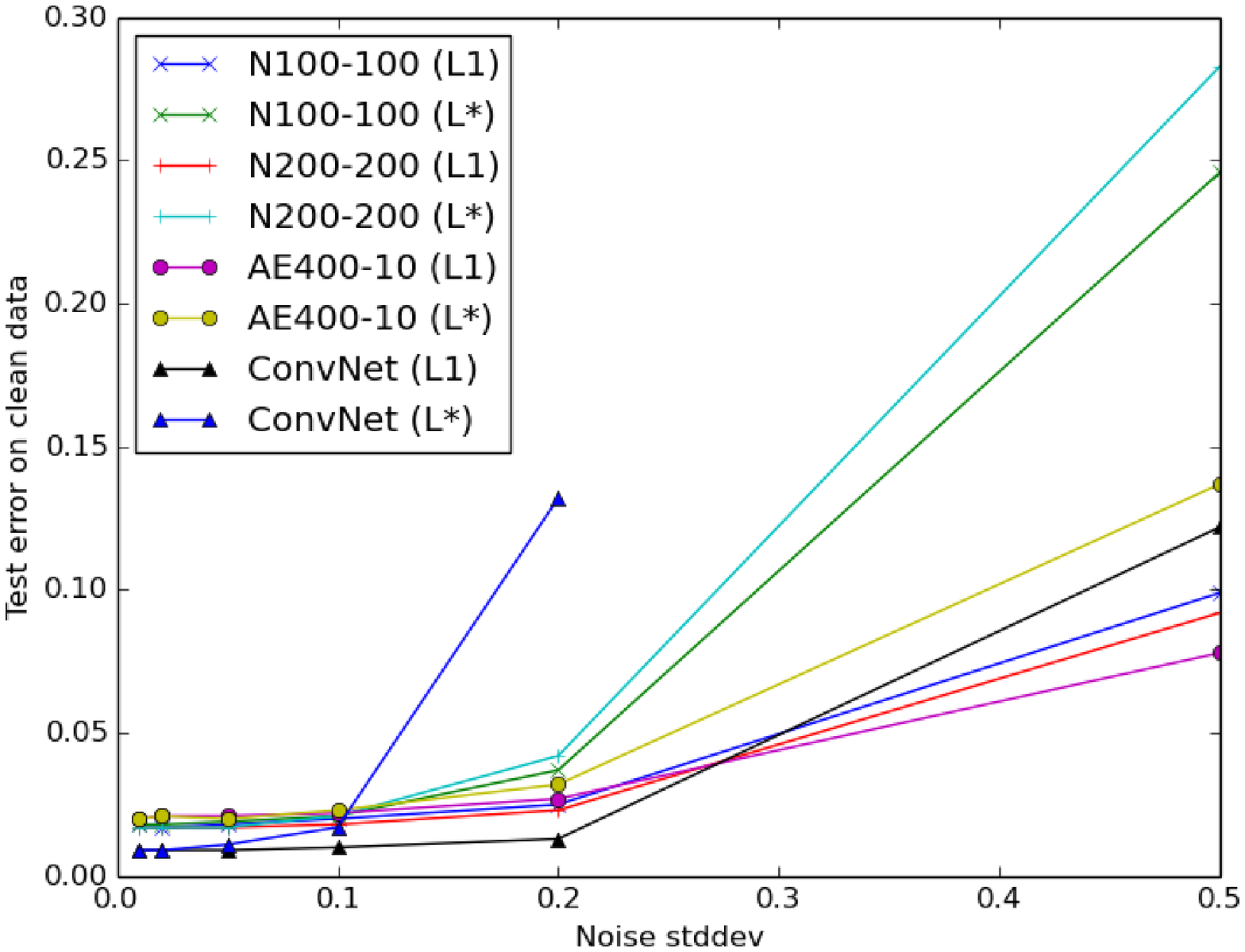}
 \end{subfigure}
 \begin{subfigure}
 \centering
  \includegraphics[width=0.49\linewidth]{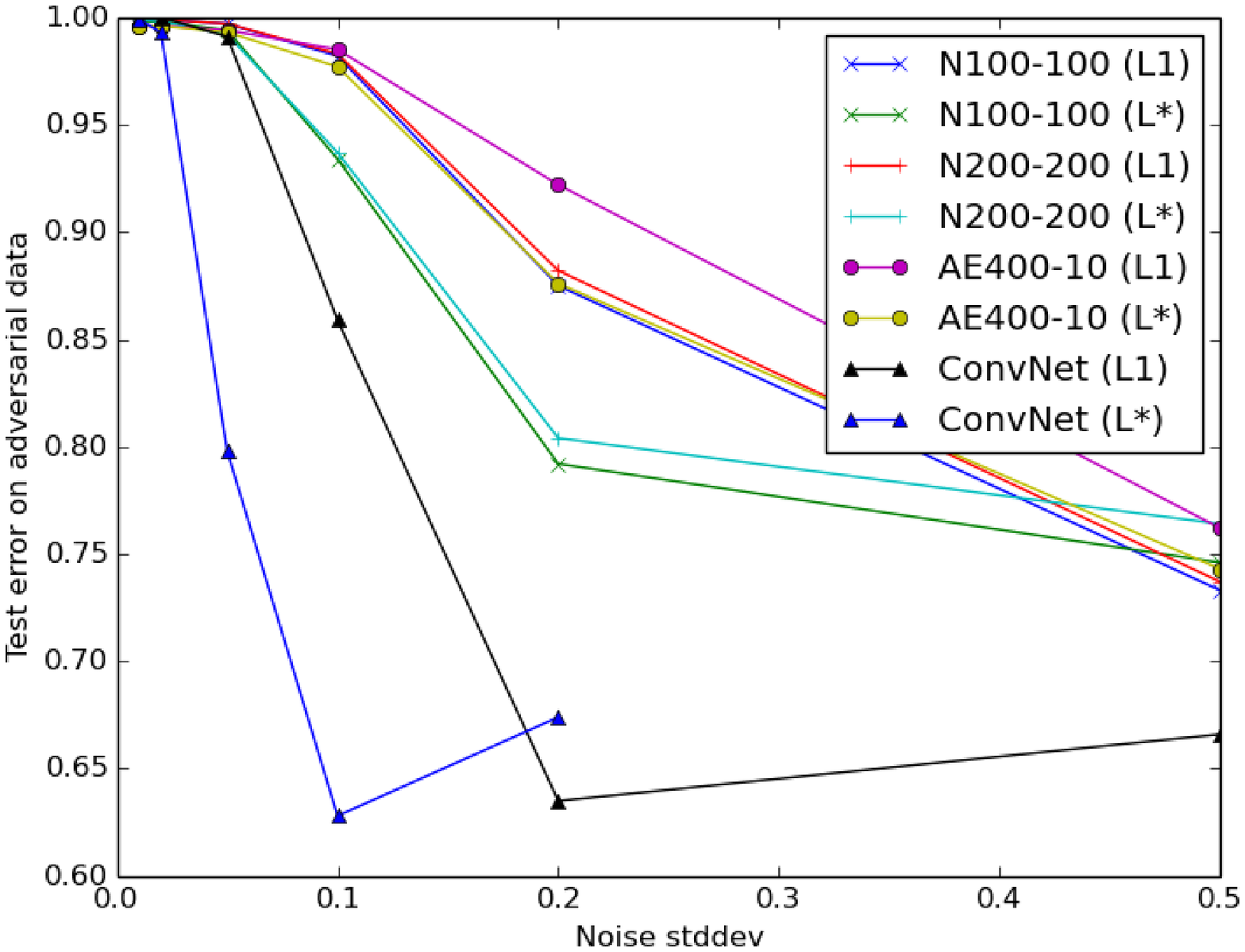}
 \end{subfigure}
 \caption{Noise Injection (Gaussian additive noise): Test error on clean data (Left) and on adversarial data (Right) vs. standard deviation of Gaussian additive noise}
 \label{fig:noise_test}
\end{figure}

Given the tiny nature of adversarial noise, we investigate a recovery strategy based on additional corruptions,
in the hope that we can move the input outside the network ``blind-spots", which we initially assumed to be small and localized. 
We experiment with additive Gaussian noise and Gaussian blurring.
For Gaussian noise we averaged predictions over 20 feed-forward runs to reduce the prediction variance. 
Results for Gaussian additive noises are summarized in Figure~\ref{fig:noise_test}. 
It shows the  and the trade-off between the adversarial examples recovered and the clean examples misclassified as one varies
the amount of additive noises, which are added to only input layer or the input plus all the hidden layers.
Results for Gaussian blurring are summarized in Table~\ref{table:blur-table}.
For Gaussian additive noises, ``L1'' refers to the noise applied at input layer; ``L*'' refers to the noise applied at input layer plus all hidden layers. 
Gaussian blurring is only applied at input layer. 

The results show that convolution seems to help with recovering from the adversarial examples.
For example, for ConvNet model, applying Gaussian blur kernel of size 11 to all input data can recover more than 50\% of adversarial examples,
at the expense of 3\% increase in the test error on clean data (Table~\ref{table:blur-table}). In addition, 
for ConvNet model, adding Gaussian noise of $\sigma=0.1$ at input layer plus hidden layers allow the model to recover
more than 35\% at similar small loss in model performance on clean data (Figure~\ref{fig:noise_test}). 
However, neither Gaussian additive noises or blurring is effective in removing enough noise such that its error on adversarial examples
could match that of the error on clean data. 

\begin{table}[t]
\caption{Noise Injection (Gaussian blurring): Test error on clean data (Left) and on adversarial data (Right) vs. blur kernel size}
\label{table:blur-table}
\begin{center}
\begin{tabular}{|l|lll|lll|}
\hline
Blur Kernel Size &- &5 &11 &- &5 &11\\ 
\hline
N100-100-10 &1.8   		&2.6	&11.3&99.9   	&43.5	&62.8\\
N200-200-10 &1.6   	&2.5	&14.8&99.9    	&47.0	&65.5\\
AE400-10    &2.0   	&3.2	&16.6 &99.6   		&68.3	&78.8\\
ConvNet	&0.9 		&1.2	&4.0&100		&53.8	&43.8\\
\hline
\end{tabular}
\end{center}
\end{table}

%
%
%

\subsection{Autoencoder}
\label{sec:autoencoder}

To assess the structure of the adversarial noise, we trained a three-hidden-layer autoencoder (784-256-128-256-784 neurons)
on mapping adversarial examples back to the original data samples.
An important detail is that we also train the model to map original training data back to itself, so that the autoencoder 
preserves the original data if the non-adversarial data samples are fed in; this allows us to stack several autoencoders.
We train the autoencoder using adversarial examples from the training set only, and test generalization capabilities on adversarial
examples from the test set across different model topologies. 
Table~\ref{ae-table} shows generalization performance of autoencoders trained on adversarial examples from different models. 
Columns indicate whose adversarial data the autoencoder is trained on, rows indicate whose adversarial test data the autoencoder
is used to denoise. Entries correspond to error rates when the outputs from the autoencoder is fed into the model identified by the row labels. 
We observe that autoencoders generalize very well on adversarial examples from different models. 
All autoencoders are able to recover at least 90\% of adversarial errors, regardless of the model from which it originates.

\begin{figure}
 \begin{center}
  \includegraphics[scale=0.30]{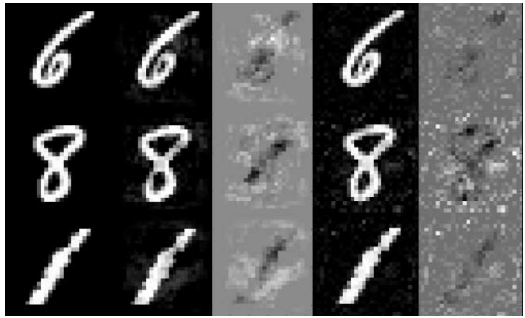}
  \caption{1st column is the original data. 2nd and 3rd are the adversarial examples and their noises for the 
  original model. 4th and 5th are for the stacked model of AE + the original net.}
 \label{fig:adv_serialized}

 \end{center}
\end{figure}

While this is a very successful experiment, we found one drawback: the autoencoder and its corresponding classifier can be stacked
to form a new feed-forward neural network, then adversarial examples can again generated from this stacked network.
The last row in Table~\ref{ae-table} and Figure~\ref{fig:adv_serialized} show such stacked network adversarial examples to have a 
significantly smaller distortion than adversarial examples from the original classifier network, suggesting that while the autoencoder
effectively recovers the from the weaknesses of the original classifier network, the stacked network is then even more susceptible to adversarial noises.
One possible explanation is that since the autoencoder is trained without the knowledge of the classification objective, 
it has more ``blind-spots'' with respect to that final objective. This again confirms the necessity of end-to-end training in deep neural networks.

\begin{table}[t]
 \caption{Cross-model autoencoder generalization test. Error-rates on adversarial test data. Last two rows
 shows the error-rates on clean test data and the average minimal distortion of adversarial examples generated from stacked network,
 respectively. The error-rates on the adversarial test data without the autoencoder preprocessing is approximately 100\% as shown in Table~\ref{models-table}.}
 \label{ae-table}
 \begin{center}
 \begin{tabular}{|l|llll|}
 \hline
   &N-100-100-10	&N200-200-10	&AE-400-10	&ConvNet\\
 \hline
 N-100-100-10	&2.3\%	&2.4\%	&2.3\%	&5.2\%	\\
 N-200-200-10	&2.3\%	&2.2\%	&2.2\%	&5.4\%	\\
 AE400-10	&3.6\%	&3.5\%	&2.7\%	&9.2\%	\\
 ConvNet	&7.7\%	&7.6\%	&8.3\%	&2.6\%	\\
 \hline
 Test error (clean)	&2.1\% 	&1.9\%	&2.1\%	&1.1\% \\
 Avg adv distortion	&0.049	&0.051	&0.043	&0.038 \\
 \hline
 \end{tabular}
 \end{center}
\end{table}

\subsection{Denoising Autoencoder}
\label{sec:dae}

In this section, a standard denoising autoencoder (DAE) is trained, without the knowledge of the adversarial noise distribution.
A DAE maps corrupted input data to clean input data. At each training batch, each pixel in the input data is corrupted by adding 
independent Gaussian noise with 0 mean and $\sigma$ standard deviation. 
Table~\ref{dae-table} summarizes the results indicating that a standard denoising autoencoder can still recover a significant portion of the adversarial noises.
In particular, a denoising auto-encoder with $\sigma=0.1$ Gaussian noise could denoise adversarial examples almost as well as an autoencoder 
trained on actual adversarials noises, as shown in Table~\ref{ae-table}. 
However, this model also suffers the same deficiency as in Section~\ref{sec:autoencoder}, that a stacked network is more susceptible to adversarials.
In this case, this deficiency likely arises due to imperfect training of DAE itself. 

\begin{table}[t]
 \caption{Denoising autoencoder test. Error on the adversarial test data.}
 \label{dae-table}
 \begin{center}
 \begin{tabular}{|l|llll|}
 \hline
    &N-100-100-10	&N200-200-10	&AE-400-10	&ConvNetM\\
 \hline
 DAE, $\sigma=0.1$	&5.0\%	&4.9\%	&11.5\%	&9.1\%	\\
 DAE, $\sigma=0.5$	&10.0\%	&10.6\%	&16.3\%	&15.3\%	\\
 \hline
 \end{tabular}
 \end{center}
\end{table}

\subsection{Discussion}
\label{noise_discuss}

Our experiments have shown that the adversarial noise is fairly robust against local perturbations such as additive Gaussian noise,
suggesting that the size of ``blind-spots'' is relatively large.
In the image data case, the effect of adversarial examples can be significantly reduced by low-pass filtering, such as Gaussian blurring, 
suggesting that adversarial noise mostly resides in high-frequency domain.
Moreover, the success of the autoencoder and the denoising autoencoder experiment shows adversarial noise to have simple structure that is easily exploitable. 

A key observation about the adversarial examples is that they are unavoidable and intrinsic property of any feed-forward architecture. 
For any pre-processing, it is always possible to backpropagate the error signal through the additional functions and find new adversarial examples, 
not only for deterministic pre-processing steps such as Gaussian blurring and autoencoders. 
Surprisingly, our experiments show that the distortion of adversarials from stacked network is even lower than
the distortion of adversarials from the original classifier network.
Furthermore, Table~\ref{dae-table} also hints that even a simple Gaussian additive noise, often used in data augmentation, effectively creates flat, 
invariant regions around the input data points. 
Based on these results and observations, we thus postulate that solving the adversarial problem should correspond to finding 
new training procedures and objective functions so as increase the distortion of the smallest adversarial examples. 
Therefore, in Section~\ref{sec:dcn_top}, we formulate a new model that could propagate the input invariance toward the final network
outputs and be trained in end-to-end. 

\section{Deep Contractive Network}
\label{sec:dcn_top}

In this section, we formulate Deep Contractive Network, which imposes a layer-wise contractive penalty in a 
feed-forward neural network. The layer-wise penalty approximately minimizes the 
network outputs variance with respect to perturbations in the inputs, enabling the trained model to 
achieve ``flatness'' around the training data points. 

\subsection{Contractive autoencoder}
\label{cae_mdae}

Contractive autoencoder (CAE) is a variant of an autoencoder (AE) with an additional penalty for
minimizing the squared norm of the Jacobian of the hidden representation with respect to input data~\cite{rifai2011contractive}. 
A standard AE consists of an encoder and a decoder. The encoder maps input data to the hidden representation,
and the decoder attempts to reconstruct the input from the hidden representation. 
Formally, given input $x\in\mathbb{R}^{d_x}$ and hidden representation $h\in\mathbb{R}^{d_h}$,
the encoder parametrized by $d_h \times d_x$ matrix $W_e$ and bias vector $b_h\in\mathbb{R}^{d_h}$ and 
the decoder parametrized by $d_x \times d_h$ matrix $W_d$ and bias vector $b_y\in\mathbb{R}^{d_h}$,
the output $y\in\mathbb{R}^{d_x}$ is given by:
\begin{equation}
 \label{eq:ae_out}
 y = \sigma_d(W_dh + b_y) = \sigma_d(W_d\sigma_e(W_ex + b_h) + b_y)
\end{equation}
where $\sigma_e$ and $\sigma_d$ are non-linear activation functions for the encoder and decoder respectively.
Given $m$ training data points, the AE is trained by finding the model parameters $\theta=\{W_e,W_d,b_h,b_y\}$
that minimize the following objective function:
\begin{equation}
 \label{eq:ae_loss}
 J_{AE}(\theta)=\sum_{i=1}^{m}\ L(x^{(i)},y^{(i)})
\end{equation}
For CAE, the objective function has an additional term:
\begin{equation}
 \label{eq:cae_loss}
 J_{CAE}(\theta)=\sum_{i=1}^{m}\ (L(x^{(i)},y^{(i)}) + \lambda\parallel\frac{\partial h^{(i)}}{\partial x^{(i)}}\parallel_2)
\end{equation}
$\lambda$ is a scaling factor that trades off reconstruction objective with contractive objective. 
$\parallel\frac{\partial h^{(i)}}{\partial x^{(i)}}\parallel_2$ is the Frobenius norm of the Jacobian matrix
of $h^{(i)}$ with respect to $x^{(i)}$.

\subsection{Deep Contractive Networks}
\label{dcn}

A Deep Contractive Network (DCN) is a generalization of the contractive autoencoder (CAE) to a feed-forward neural network
that outputs $y\in\mathbb{R}^{d_y}$ with a target $t\in\mathbb{R}^{d_y}$. 
For a network with $H$ hidden layers, let $f_i$ denote the function for computing hidden representation 
$h_i\in\mathbb{R}^{d_{h_i}}$ at hidden layer $i$: $h_i = f_i(h_{i-1})$, $i=1...H+1$, $h_0 = x$ and 
$h_{H+1}=y$. Ideally, the model should penalize the following objective:
\begin{equation}
 \label{eq:dcn_loss}
 J_{DCN}(\theta)=\sum_{i=1}^{m}\ (L(t^{(i)},y^{(i)}) + \lambda\parallel\frac{\partial y^{(i)}}{\partial x^{(i)}}\parallel_2)
\end{equation}
However, such a penalty is computationally expensive for calculating partial derivatives at each layer
in the standard back-propagation framework. Therefore, a simplification is made by approximating the objective with the following:
\begin{equation}
 \label{eq:dcn_loss_2}
 J_{DCN}(\theta)=\sum_{i=1}^{m}\ (L(t^{(i)},y^{(i)}) + \sum_{j=1}^{H+1}\lambda_j\parallel\frac{\partial h_j^{(i)}}{\partial h_{j-1}^{(i)}}\parallel_2)
\end{equation}
This layer-wise contractive penalty enables partial derivatives to be computed in the same way as in a contractive autoencoder,
and is easily incorporated into the backpropagation procedure.
This objective does not guarantee global optimality for the solution to Eq.~\ref{eq:dcn_loss}, and also limits the capacity
of the neural network. However, it is a computationally efficient way to greedily propagate input invariance through
a deep network.

Note that for a neural network with additive Gaussian noise $\mathcal{N}(0,\sigma^2I)$ added to input during the training, 
if we let $\sigma\rightarrow0$, the training objective is equivalent to Eq.~\ref{eq:dcn_loss}~\cite{alain2012regularized}.
However, such stochastic penalties require many passes of data to train the model effectively. For efficiency,
we decided to employ a deterministic penalty instead~\cite{rifai2011contractive,alain2012regularized,chen2014marginalized}.

\subsection{Experiments and results}
\label{experiments}

The experiments involve applying a contractive penalty to the models in Table~\ref{models-table}.
The models were trained until they achieved nearly the same accuracy as the original models
which lacked a contractive penalty. The adversarial examples are generated following the method defined in Section~\ref{gen_adv}. 

Table~\ref{dcn-table-1} shows that the contractive penalty successfully increases the minimum distortion of the
adversarial noises. Table~\ref{dcn-table-2} shows the comparison of the Deep Contractive Network penalty against
stochastic noise addition. Deep Contractive Networks are more robust than a standard neural network trained with
Gaussian input noise, and can be easily augmented by adding Gaussian input noise to further
increase the minimum distortion of adversarial noises. 

\begin{table}[t]
\caption{The error-rates on clean test data and the average distortion of adversarial examples generated from the original model (orig) 
and the same model with contractive penalty (DCN). The error-rates on the adversarial examples are 100\%.}
\label{dcn-table-1}
\begin{center}
\begin{tabular}{| l |l | l |l|l|}
\hline
Model Name &DCN error &DCN adv. distortion &orig. error &orig. adv. distortion\\
\hline  
N100-100-10   &2.3\% &\bf{0.107} &1.8\% &0.084\\
N200-200-10   &2.0\% &\bf{0.102} &1.6\% &0.087\\
AE400-10      &2.0\% &\bf{0.106} &2.0\% &0.098\\
ConvNet	      &1.2\% &\bf{0.106} &0.9\% &0.095\\
\hline
\end{tabular}

\end{center}
\end{table}

\begin{table}[t]
\caption{The error-rates on clean test data and the average distortion on adversarial examples from N200-200-10 models with different
training conditions. ``GN'' refers to Gaussian additive noise during training.}
\label{dcn-table-2}
\begin{center}
\begin{tabular}{|l | l |l|}
\hline
Training Condition  &Test error &Av. distortion \\
\hline  
DCN&2.0\%&0.102 \\
GN,L1,$\sigma=0.1$&1.8\%&0.095 \\
GN,L*,$\sigma=0.1$&2.0\%&0.099	\\
DCN+GN,L1,$\sigma=0.1$&2.2\%&0.108	\\
\hline
\end{tabular}
\end{center}
\end{table}

\begin{figure}
 \begin{center}
  \includegraphics[scale=0.30]{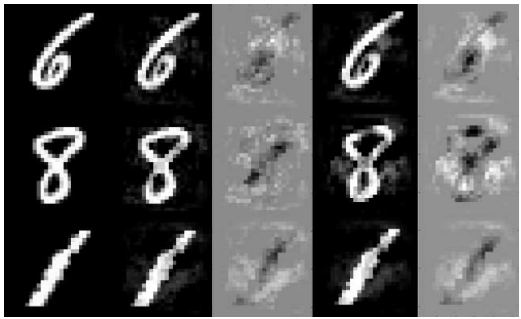}
  \caption{1st column is the original data. 2nd and 3rd are the adversarials and noises for the  original model. 
  4th and 5th are the adversarials and noises for the model trained with contractive penalty.}
 \label{fig:adv_dcn}
 \end{center}
\end{figure}

\subsection{Discussions and future work}
\label{sec:dcn_disc_fut}

Results show that Deep Contractive Networks can successfully be trained to propagate contractivity around the input data through the deep architecture,
without significant loss in final accuracies. 
The model can be improved by augmenting the layer-wise contractive penalty based on Higher-Order Contractive autoencoders~\cite{rifai2011higher}, 
and marginalized Denoising autoencoders~\cite{chen2014marginalized}. 
While in this paper, Deep Contractive Networks are used as a framework for alleviating effects due to the adversarial examples,
they also provide a suitable framework for probing the invariance properties learnt by deep neural networks. 
A further study should be conducted to evaluate the performance loss due to layer-wise penalties as opposed to global 
contractive objectives as defined in Eq.~\ref{eq:dcn_loss}. 
In addition, exploring non-Euclidean adversarial examples, e.g. small affine transformation on images, and varying contractivity 
at higher layers of the network could lead to insights 
into semantic attributes of features learned at high levels of representation. For example, explicitly learning instantiation parameters as previously attempted
by models such as Transforming Autoencoder~\cite{hinton2011transforming}. 

Our work also bridges supervised learning with unsupervised representation learning, by introducing the penalty from DAE and CAE to standard DNN. 
Such penalty not only acts as practical regularizers, but also is a highly efficient way to learn obvious information from the training data,
such as local generalization in the input space. 
Recent progress in deep neural networks is driven by both end-to-end supervised training and various modes of 
unsupervised feature learning~\cite{krizhevsky2012imagenet,bengio2009learning},
and thus we believe the merge of the two could likely enable new milestones in the field. 

\section{Conclusions}
\label{conclusions}

We tested several denoising architectures to reduce the effects of the adversarial examples, and conclude that while the simple and stable structure 
of adversarial examples makes them easy to remove with autoencoders, 
the resulting stacked network is even more sensitive to new adversarial examples.
We conclude that neural network's sensitivity to adversarial examples is more related to intrinsic deficiencies in the training procedure
and objective function than to model topology.
The crux of the problem is then to come up with an appropriate training procedure and objective function that can efficiently make the 
network learn flat, invariant regions around the training data. 
We propose Deep Contractive Networks to explicitly learn invariant features at each layer and show some positive initial results.

\subsubsection*{Acknowledgments}

The authors thank Nitish Srivastava for releasing his
DeepNet Library, and Volodymyr Mnih for his CUDAMat Library~\cite{mnih2009cudamat}.

\bibliography{iclr2015}
\bibliographystyle{iclr2015}

\end{document}